\documentclass[sigconf]{acmart}
\copyrightyear{2021}
\acmYear{2021}
\pdfoutput=1
\setcopyright{acmcopyright}\acmConference[ICMI '21]{Proceedings of the 2021
International Conference on Multimodal Interaction}{October 18--22, 2021}{Montréal,
QC, Canada}
\acmBooktitle{Proceedings of the 2021 International Conference on Multimodal
Interaction (ICMI '21), October 18--22, 2021, Montréal, QC, Canada}
\acmPrice{15.00}
\acmDOI{10.1145/3462244.3479901}
\acmISBN{978-1-4503-8481-0/21/10}

\usepackage{amsmath}
\usepackage[gen]{eurosym}
\usepackage{hyperref}
\usepackage{enumitem}
\setlist[itemize]{noitemsep, topsep=0pt}
\usepackage{graphicx}
\usepackage{caption}
\usepackage{comment}
\usepackage{subcaption}
\usepackage[utf8]{inputenc}
\usepackage[linesnumbered,ruled,vlined]{algorithm2e}
\usepackage{multirow}
\usepackage{subcaption}
\AtBeginDocument{%
  \providecommand\BibTeX{{%
    \normalfont B\kern-0.5em{\scshape i\kern-0.25em b}\kern-0.8em\TeX}}}
\usepackage{balance }
\setcopyright{acmcopyright}

\begin{document}
\def\x{{\mathbf x}}
\def\L{{\cal L}}
\def\eg{\textit{e.g.}}
\def\ie{\textit{i.e.}}
\def\Eg{\textit{E.g.}}
\def\etal{\textit{et al.}}
\def\etc{\textit{etc}}
\newcommand{\blue}[1]{\textcolor{blue}{#1}}

\title{\emph{Head Matters}: Explainable Human-centered Trait Prediction from Head Motion Dynamics}

%
\author{Surbhi Madan}
\affiliation{Indian Institute of Technology (IIT), Ropar}
\email{surbhi.19csz0011@iitrpr.ac.in}

\author{Monika Gahalawat}
\affiliation{Indian Institute of Technology (IIT), Ropar}
\email{monika.20csz0003@iitrpr.ac.in}

\author{Tanaya Guha}
\affiliation{University of Warwick}
\email{tanaya.guha@warwick.ac.uk}

\author{Ramanathan Subramanian}
\affiliation{IIT, Ropar and University of Canberra}

\email{ram.subramanian@canberra.edu.au}
%

\renewcommand{\shortauthors}{ICMI '21, October 18--22, 2021, Montréal,
QC, Canada}

\begin{abstract}
We demonstrate the utility of elementary head-motion units termed \emph{\textbf{kinemes}} for behavioral analytics to predict \emph{personality} and \emph{interview} traits. Transforming head-motion patterns into a sequence of kinemes facilitates discovery of latent temporal signatures characterizing the targeted traits, thereby enabling both \emph{efficient} and \emph{explainable} trait prediction. Utilizing Kinemes and Facial Action Coding System (FACS) features to predict (a) OCEAN personality traits on the First Impressions Candidate Screening videos, and (b) Interview traits on the MIT dataset, we note that: (1) A Long-Short Term Memory (LSTM) network trained with kineme sequences performs better than or similar to a Convolutional Neural Network (CNN) trained with facial images; (2) Accurate predictions and explanations are achieved on combining FACS action units (AUs) with kinemes, and (3) Prediction performance is affected by the time-length over which head and facial movements are observed.       
\end{abstract}
\vspace{-3mm}

\vspace{-2mm}


\maketitle
\vspace{-3.5mm}
\section{Introduction}\label{Sec:Intro}
The importance of {non-verbal} behavioral cues towards human-centric trait estimation has been acknowledged for long. Case in point, multifarious cues like proxemics or the use of physical space~\cite{Takayama09}, speaking time and speech energy~\cite{Jayagopi09}, head orientation patterns denoting social attention distribution~\cite{Subramanian13,Subramanian10}, head and facial movements~\cite{escalante2020modeling,Gucluturk2018} and physiological responses to emotional stimuli~\cite{Subramanian18} have been studied for predicting {personality traits} driving human behavior. While a majority of these cues are relevant and effectively captured in social settings, there has been increasing interest in predicting human-centered traits from self-presentation videos (alternatively termed \emph{multimedia CVs})~\cite{Batrinca11} recently. 

\begin{figure}[tb]
    \centering
    \includegraphics[width=\linewidth, trim= {0cm 0cm 0cm 0cm},clip=true]{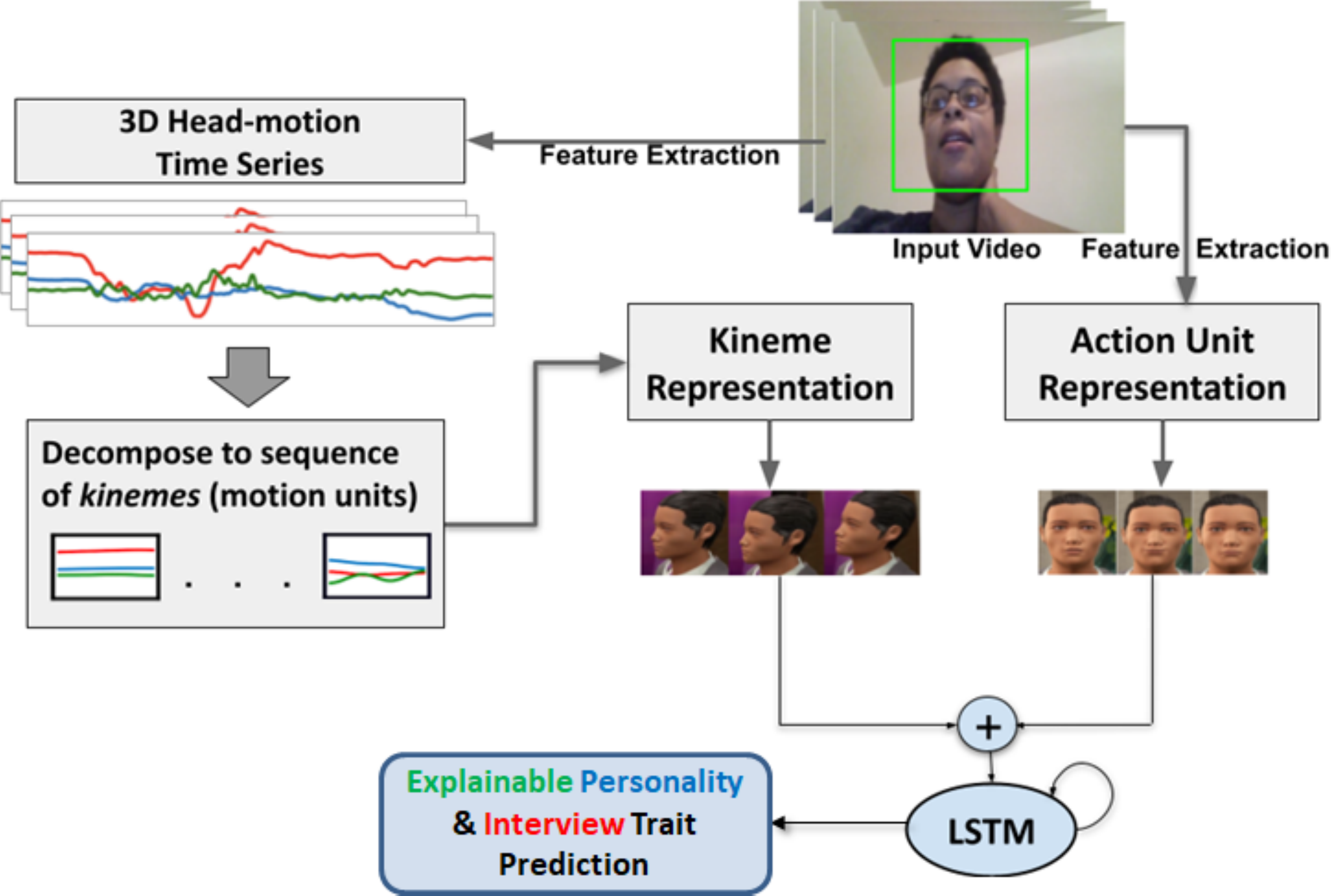}\vspace{-4mm}
    \caption{Framework overview. Kinemes and action units are depicted via a 3D model. View in color and under zoom.}\label{fig:teaser}\vspace{-6.2mm}
\end{figure}

Multimodal non-verbal cues especially play a critical role in demonstrating an individual's personality and inter-personal skills in the context of multimedia CVs~\cite{Raza1987AMO,Batrinca11}. Subjective impressions of interviewee personality traits influence hiring decisions~\cite{VanDam03}, and even one behavioral information channel such as \emph{visual} can explain personality attributions~\cite{DeGroot09}. \emph{E.g.}, among the big-five or OCEAN personality traits~\cite{Chmielewski2013}, \emph{Conscientiousness} characterizing responsible and attentive behavior is reflected via an upright posture and minimal head movements in self-presentation videos. Likewise, \emph{Neuroticism} indicating anxiety and stress is revealed via hand movement and posture dynamics such as fidgeting and camera aversion~\cite{Batrinca11}.    

The above examples convey that head-motion plays a critical role in conveying personality impressions. Recently, \emph{kinemes} denoting elementary head-motion units akin to phonemes in human speech~\cite{birdwhistell1970}, are shown to be effective emotional cues~\cite{samanta2021emotion}. Apart from enabling human-comparable recognition performance, kinemes also allow for \emph{explanations} relating to predictions; \eg, the sad emotion is associated with slow head movements~\cite{gross2010}.

This work explores the utility of kinemes for predicting \emph{personality} and \emph{interview} traits from multimedia CVs. As in Figure~\ref{fig:teaser}, we extract head-motion from an input video via the 3D \emph{yaw}, \emph{pitch} and \emph{roll} Euler rotation angles. The head motion time-series is then unsupervisedly decomposed into a kineme sequence comprising elementary head motion units. This decomposition enables discovery of kinemes characteristic of a target trait, \eg, head nodding is typical of courteous or sympathetic behavior~\cite{Ishii2020,Kawahara18}. In addition to head motion, we also utilize action units (AUs) which encode facial motion and expressions, for trait prediction. Our experiments confirm that encoding head and facial movements in terms of kinemes and AUs enables efficient and explanative trait predictions. Overall, we make the following research contributions:
\begin{itemize}
\item[(1)] We novelly employ kinemes for personality and interview trait prediction. While medium-grained head motion behaviors like social gaze have been shown to predict traits like Extraversion~\cite{Lepri12}, we show that kinemes denoting fine-grained and elementary head motion are effective non-verbal behavioral cues; Kinemes are also interpretable, can be learned without supervision from input videos, and can enable discovery of latent temporal signatures of the target trait. 
\item[(2)] Our experiments reveal that kinemes are highly predictive of personality and interview traits. A kineme-based long short term memory (LSTM) network performs comparable to a 2D-convolutional neural network (CNN) for personality trait prediction on the First Impressions Candidate Screening (FICS) dataset constituting 10K videos. On the 138-videos MIT dataset annotated for interview-specific traits, the kineme LSTM considerably outperforms 2D-CNN. Fusing kinemes with AUs also improves prediction performance owing to the complementary information they encode.
\item[(3)] Apart from being predictive, kinemes and AUs enable behavioral explanations for the target traits. As examples, frequent head nodding and smiling convey high Agreeableness, while head shaking and frowning are indicative of low Agreeableness; likewise, upward head-tilt indicating upright demeanor conveys high Conscientiousness; conversely, looking down to avoid eye-contact is indicative of low Conscientiousness.
\item[(4)] We present empirical results and ablative studies on two diverse datasets, namely the FICS and MIT interview datasets. With both kinemes and AUs, we examine personality and interview trait prediction from very short behavioral episodes known as \emph{thin-slices}. While reasonable prediction performance is achieved even with 5s-long slices, more precise predictions are achieved with longer behavioral slices. Also, we assess the suitability of video labels to thin-slices by comparing slice and video-level predictions; video labels are consistent with longer-slice behaviors for both datasets. \vspace{-2mm}
\end{itemize}

\section{Related Work}\label{Sec:RW}

This section reviews literature on (a) personality and interview trait prediction, and (b) employing head motion features for behavioral analytics, and highlighting the research gaps thereof. 

\subsection{Personality and Interview Trait Prediction}
It is well known that personality drives human behavior, and the big-five or OCEAN personality trait model~\cite{Costa1992} describes human personality in terms of the Openness (curious vs cautious), Conscientiousness (diligent vs insincere), Extraversion (outgoing vs reserved), Agreeableness (sympathetic vs dispassionate) and Neuroticism (nervous vs emotionally stable) dimensions. Numerous studies have attempted personality prediction from behavioral, and specifically non-verbal cues such as proxemics~\cite{Takayama09}, speaking and head movement behavior~\cite{Jayagopi09,Subramanian13}, facial characteristics~\cite{escalante2020modeling,Gucluturk2018} and physiological responses to emotional scenes~\cite{Subramanian18}. 

Many studies have also examined the relationship between personality traits and job/job-interview performance. \emph{E.g.}, Mount~\etal~\cite{Murray1998} observe that Conscientiousness, Agreeableness and Emotional stability positively impact job performance involving interpersonal interactions. Authors of~\cite{Moy03} observe that the OCEAN traits are among the major attributes influencing hiring decisions. Two recent studies that examine the relation between personality traits and human factors in job interviews are~\cite{Gucluturk2018,escalante2020modeling}. Naim~\etal~\cite{Naim18} perform multimodal analyses of interview videos and conclude that prosodic features (speaking style) critically influence impressions of interview-specific traits.\\

\noindent \textbf{Explainable personality and interview trait prediction:} While deep learning architectures such as CNNs and LSTMs have achieved excellent performance on multiple pattern recognition problems, their predictions are often not interpretable~\cite{Ventura17}. A deep residual network (ResNet) is proposed in~\cite{Gucluturk2018} to predict personality trait impressions, and linear regression is employed to predict interview scores from personality trait annotations; regression coefficients are employed to assess the influence of the OCEAN traits on interview scores. Face visualizations are also presented to demonstrate similarities among individuals achieving high and low trait scores. CNN-based trait prediction is explored in~\cite{Ventura17}, and analyses show that CNNs primarily analyze key facial regions such as eyes and mouth for prediction. The FICS dataset developed for a candidate screening challenge is presented in~\cite{escalante2020modeling}. Decision trees and visual-plus-verbal explanations are presented to convey relationships among personality and interview scores. Facial action units, which are typically used to describe emotions are shown to effectively predict personality traits in~\cite{Gavrilescu15}, and the correlations among AUs and learned CNN features is demonstrated in~\cite{Ventura17}. \vspace{-2mm}

\subsection{Head Motion for Behavioral Analytics} A majority of existing works on head-motion rely on extracting low-level features from head motion data. For example, Ding~\etal~\cite{ding2018low} use amplitude of representative Fourier components. Samanta and Guha~\cite{samanta2017role} propose to extract energy of displacement, velocity and acceleration of the Euler rotation angles, while Gunes and Pantic~\cite{gunes2010dimensional} employ magnitude and direction of the 2D head motion. The use of semantically meaningful head gestures has been limited to extracting nods and shakes~\cite{gunes2010dimensional}. However, head motion generated during dyadic interactions and self-presentation videos is complex; therefore high-level head gestures may not be limited to only nods and shakes. To this end, Yang and Narayanan~\cite{yang2017modeling} propose to extract arbitrary head gesture segments. These head gesture segments are abstract, and do not have physical interpretation. \\

\noindent\textbf{Head motion in human-centered traits:} In the context of affect analysis, head motion patterns have been used to study coordination between mothers and infants~\cite{hammal2015head,hammal2015can}, emotion recognition~\cite{samanta2021emotion}, measuring engagement levels of dementia patients~\cite{Parekh18} and for analyzing interpersonal coordination in couple therapy \cite{hammal2014interpersonal, xiao2015head}. 

\subsection{Inference Summary}
Analysis of the literature reveals the following shortcomings: (1) While many works examine prediction of personality and interview traits, and explain predictions via statistical analysis or visualizations, these explanations are limited to discovering salient facial features or examining connections between interview performance and personality traits. (2) While head motion patterns have been identified as critical non-verbal behavioral cues in interactive scenarios, they have nevertheless not been employed for personality or interview trait prediction.

Differently, we novelly attempt explanations for personality and interview traits from kinemes which are inherently explanatory; apart from generic explanations such as head-nodding evoking positive and head-shaking eliciting negative impressions, trait-specific characteristics such as Openness and emotional stability associating with persisting head movements, and Conscientiousness being impacted by head tilting to maintain/avoid eye-contact are also evident. More intuitive explanations are achieved on combining kinemes and facial action units such as nodding and expressive facial behavior achieving high interview scores, and head shaking and frowning being seen as less friendly and sociable. In addition to their explanatory characteristics, kinemes and AUs are also found to effectively predict the targeted traits.

\section{Kineme Formulation}\label{Sec:KF}

Given a self-presentation video, we extract the 3D head pose in terms of Euler rotation angles about $x, y, z$, namely, \emph{pitch} ($\theta_p$), \emph{yaw} ($\theta_y$) and \emph{roll} ($\theta_r$), for each frame using the computer vision tool OpenFace~\cite{Baltrusaitis16}. Thus, head motion is denoted as a multivariate time-series of head orientations: $\boldsymbol{\theta} = \{\theta_p^{1:T}, \theta_y^{1:T}, \theta_r^{1:T}\}$ of length $T$. We propose to model head motion as a sequence of fundamental and \emph{interpretable} motion units termed kinemes, which consequently enable \emph{explanative} trait predictions. Past work on motion modeling has focused on extracting head-motion patterns such as nods and shakes~\cite{gunes2010dimensional}, or learning arbitrary head gestures~\cite{yang2017modeling} with no physical meaning. Differently, we unsupervisedly learn meaningful motion patterns following~\cite{samanta2021emotion} to translate head motion into a sequence of kinemes.

%


\subsection{Head motion as kineme sequence}\label{HMtoKin}
Given the head-motion time-series extracted from a set of videos, we divide each time-series $\boldsymbol{\theta}$ into short overlapping segments of length $\ell$ (overlapping segments enable shift-invariance, are empirically found to generate better representations in~\cite{samanta2021emotion}). The $i^{th}$ segment is denoted by a vector $\mathbf{h}^{(i)} = [\theta_p^{i:i+\ell}\, \theta_y^{i:i+\ell}\, \theta_r^{i:i+\ell}]^\intercal$. The characterization matrix $\mathbf{H}_{\boldsymbol\theta}$ is defined as 
$ \mathbf{H}_{\boldsymbol\theta} = [\mathbf{h}^{(1)}, \mathbf{h}^{(2)},\cdots, \mathbf{h}^{(s)}]$, where $s$ is the total number of segments in $\boldsymbol{\theta}$. Given $N$ training samples, a head motion matrix $\mathbf{H}\in\mathbb{R}_+^{3\ell\times Ns}$ is created as $\mathbf{H} = [\mathbf{H}_{\boldsymbol\theta_1}|\mathbf{H}_{\boldsymbol\theta_2}|\cdots|\mathbf{H}_{\boldsymbol\theta_N}]$. $\mathbf{H}$ is then subject to a Non-negative Matrix Factorization (NMF), yielding a basis matrix $\mathbf{B}$ and a coefficient matrix $\mathbf{C}$. We cluster head motion segments in the transformed space by grouping the coefficient vectors (columns in $\mathbf{C}$) via a Gaussian Mixture Model (GMM) into $k<<Ns$ clusters. This produces a $k$ column matrix ${\mathbf{C}^*}$. Kinemes are transformed back to the original \emph{yaw}-\emph{pitch}-\emph{roll} space via $\mathbf{H}^*=\mathbf{B}\mathbf{C}^*$, whose columns yield the set of $k$ kinemes $\mathcal{K}$.

Upon learning kinemes for the input video set, we can represent any head motion time-series as a sequence of kinemes by associating each $\ell$-long segment to one of the $K$ kinemes. Consider the $i^{th}$ segment in $\boldsymbol{\theta}$ characterized by $\mathbf{h}^{(i)}$. We project $\mathbf{h}^{(i)}$ onto the learned subspace to obtain $\mathbf{c}^{(i)}$:
\begin{equation*}
    \hat{\mathbf{c}} = \underset{\mathbf{c}^{(i)} \geq 0}{\text{arg min}} \lVert{\mathbf{h}^{(i)} - \mathbf{B}\mathbf{c}^{(i)}}\rVert_F^2
\end{equation*}
The corresponding kineme $K^{(i)}$ for the $i^{th}$ segment is given by maximizing the posterior probability $P({K}|\hat{\mathbf{c}})$, over all $K\in\mathcal{K}$. On mapping each head motion time-series segment to a kineme, we get the corresponding kineme sequence $\boldsymbol\theta: \{K^{(1)} \cdots K^{(s)}\}, K^{(j)}\in \mathcal{K}$.

\subsection{Trait prediction from kinemes}
Prior studies have shown that human-centered traits are characterized by specific non-verbal (and specifically, head-motion) behaviors; \eg, an upright posture maintaining eye-contact conveys high Conscientiousness, while gaze avoidance is indicative of low Conscientiousness~\cite{Hoppe18,researchdigest220714}. Likewise, frequent head nodding is seen as courteous and agreeable behavior, while head shaking and frowning indicates a cold demeanor~\cite{Ishii2020,Kawahara18}. Kinemes inherently enable discovery of temporal head-motion patterns characteristic of a given trait, and sequence learning methods such as Hidden Markov Models (HMMs) and Long-short term memory (LSTM) networks can be employed to learn these latent temporal signatures for continuous or categorical trait prediction.

\section{Explainable Trait Prediction}\label{Sec:UC}

\begin{figure}[!htbp]
      \centering
     \includegraphics[width=\linewidth]{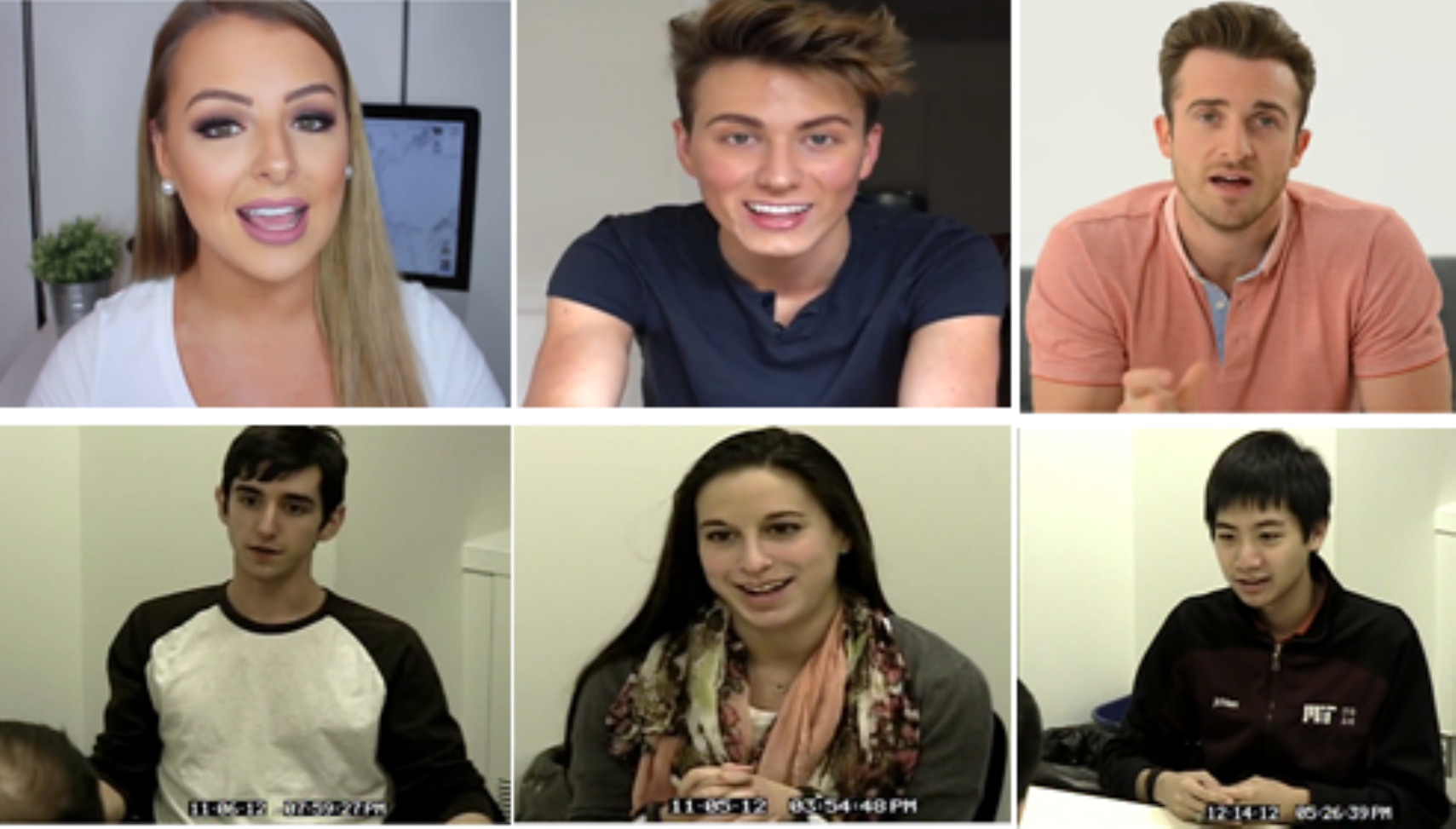}\vspace{-2mm}
    \caption{FICS (top) and MIT (bottom) video examples.} \label{fig:Ov_FICS_MIT}\vspace{-4mm}
\end{figure}

\begin{figure}[!htbp]
      \centering
     \includegraphics[width=\linewidth]{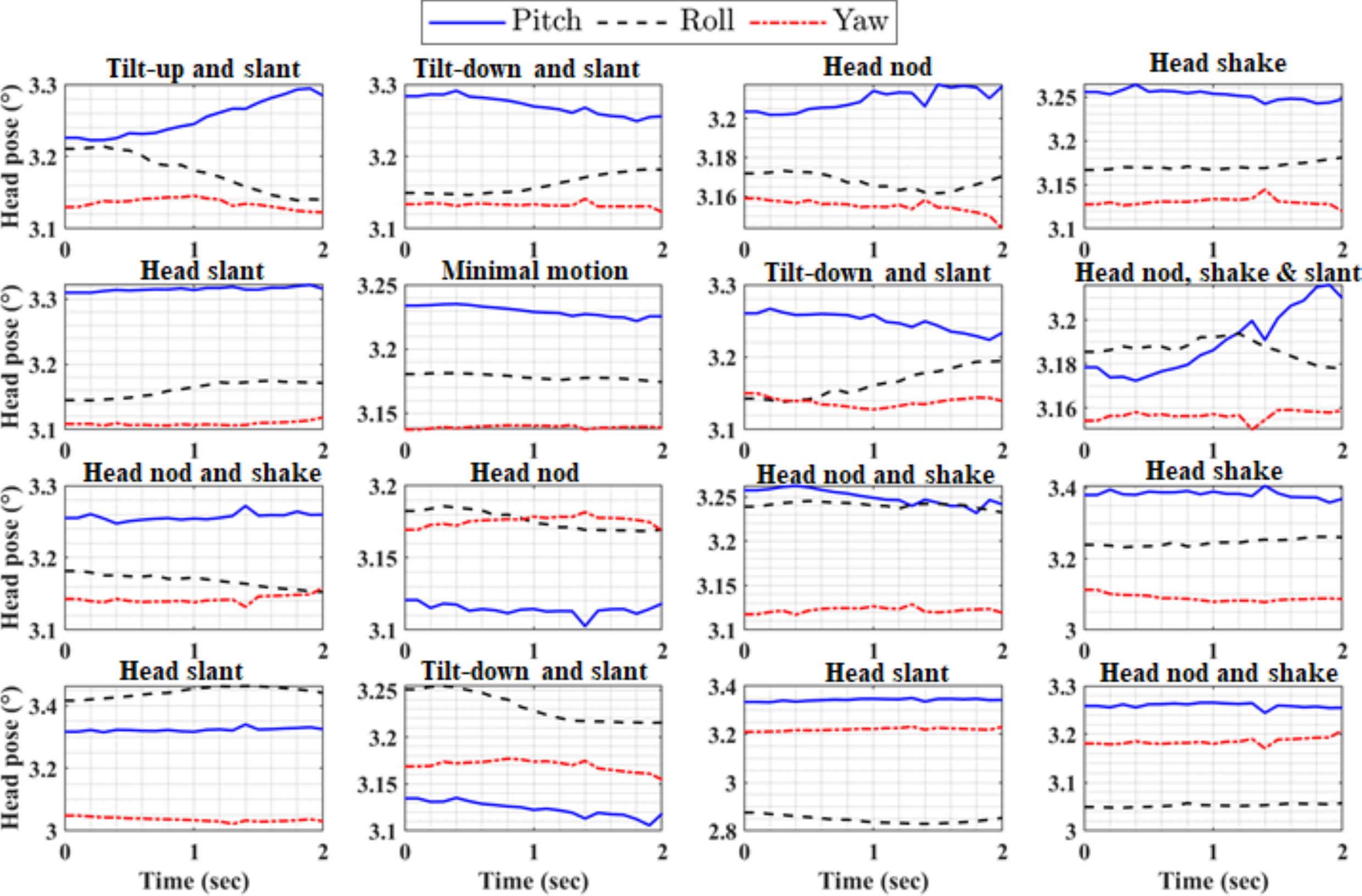}\vspace{-2mm}
    \caption{Plots of 16 kinemes extracted for the FICS dataset following raster ordering (left to right, top to bottom.} \label{fig:selectKinemes_FICS}
      \centering
     \includegraphics[width=\linewidth,height=2.3cm]{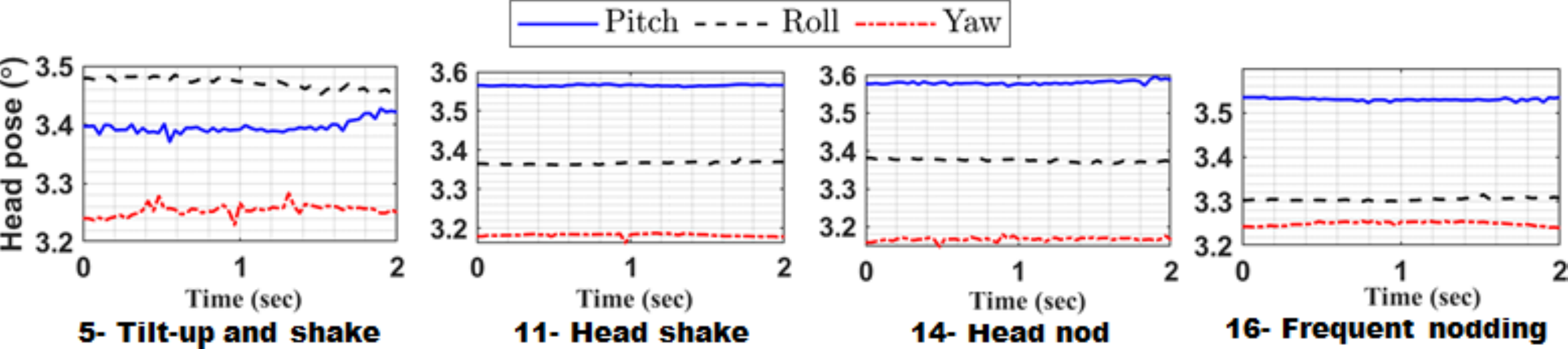}\vspace{-2mm}
    \caption{Selected kineme plots for the MIT dataset.} \label{fig:selectKinemes_MIT}
\end{figure}

\begin{figure}[!htbp]
     \includegraphics[width=\linewidth]{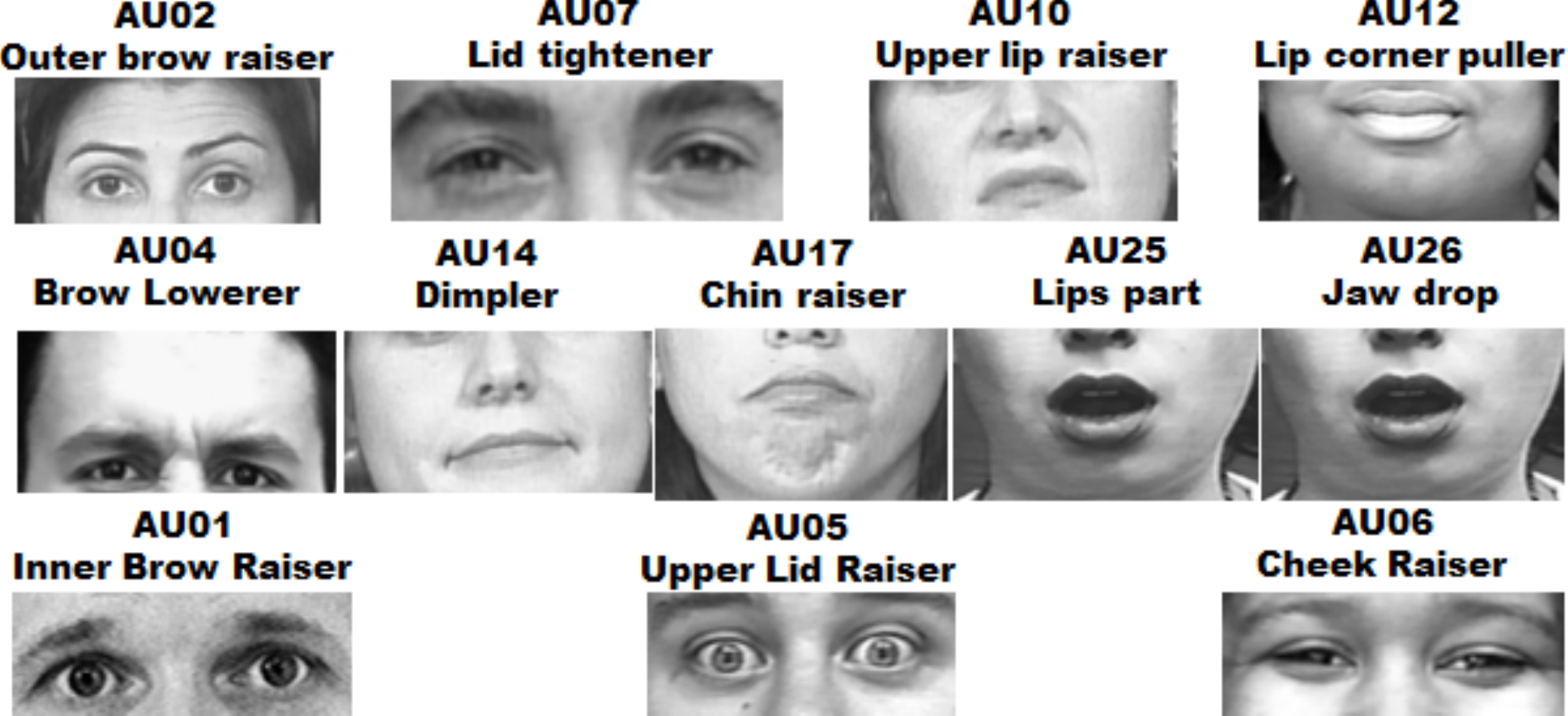}\vspace{-2mm}
    \caption{Common AUs in the FICS and MIT datasets.} \label{fig:selectAUs}
\end{figure}

In this work, we examined the First Impression Candidate Screening (FICS)~\cite{escalante2020modeling} and the MIT interview~\cite{Naim18} datasets for trait prediction. A detailed description of these datasets is presented in Sec.~\ref{sec:datasets}. The FICS dataset is curated with the objective of training algorithms to predict apparent personality traits from multimedia CVs, and has annotations for the OCEAN traits on a $[0,1]$ scale. The MIT dataset comprises recordings of mock interviews with prospective interns, and observer ratings for 16 interviewee-specific traits. Among them, we examined the following traits, hypothesizing that they could be adequately explained by head and facial behaviors: level of friendliness (Fr), excitement (Ex) and eye-contact (EC), and recommended hiring score (RH) conveying the likelihood of the candidate being invited for further interviews. Figure~\ref{fig:Ov_FICS_MIT} presents exemplar frames from the FICS and MIT datasets. While the FICS self-presentation videos are recorded under varied conditions such as differing scene background, camera perspective and camera distance from subject, the MIT videos are captured with a relatively stable background via two wall-mounted cameras.  
Upon extracting the \emph{yaw}, \emph{pitch} and \emph{roll} angles and 17 facial action unit (AU) intensities per frame with the \emph{Openface}~\cite{Baltrusaitis16} toolkit, we computed kinemes for the FICS and MIT datasets with $K = 16$ as per the procedure outlined in Sec.~\ref{HMtoKin}. We employed 2s-long segments with 50\% overlap for kineme extraction. Figure~\ref{fig:selectKinemes_FICS} presents the 16 kinemes extracted for the FICS dataset, while Fig.~\ref{fig:selectKinemes_MIT} shows selected kinemes for the MIT data. A head nod corresponds to a sudden change (spike) in pitch or head-tilt, while a head shake translates to a spike in yaw. To discover dominant AUs, we again examined 2s time-windows with 1s overlap and regarded an AU as dominant if within the window its maximum value exceeded the mean intensity over all AUs. Fig.~\ref{fig:selectAUs} presents commonly seen AUs for both datasets.

\begin{table*}[!htbp]
    \centering
		\small
    \caption{Explaining OCEAN and interview traits via kinemes and AUs. MIT Kinemes in bold font are visualized in Figure~\ref{fig:selectKinemes_MIT}.} \vspace{-2mm}
    \begin{tabular}{c|l|c|c|l}
    \toprule
     \bf Dataset & \bf Trait  & \bf Dominant Kin & \bf Dominant AUs & \bf Inferences \\
     \hline
      \multirow{10}{*}{\textbf{FICS}} & \textbf{O (H)} & 2, 8, 10, 16 &  7, 12, 14, 25, 26 & Persistent head movements (as noted in~\cite{KOPPENSTEINER2013}) with nodding and smiling.\\
      &\textbf{C (H)} & 1, 8, 10, 16 & 7, 12, 17, 25, 26 & Upward head-tilt indicative of upright demeanor and head nodding. \\
      &\textbf{E (H)} & 2, 10, 14, 16 & 10, 12, 17, 25, 26 & Head tilt-down with nodding, and facial gestures related to speaking.\\
      &\textbf{A (H)} & 3, 8, 10, 16 & 7, 12, 14, 25, 26 & Frequent head nodding and smiling (associated with courteous behavior ~\cite{Ishii2020,Kawahara18}).\\ 
      &\textbf{N (H)} & 2, 8, 10, 16 & 7, 12, 17, 25, 26 & Frequent head movements with nodding and smiling.\\ 
			&\textbf{O (L)} & 1, 6, 11, 16 & 4, 10, 14, 17, 26 & Relatively fewer head movements and frowning. \\
			&\textbf{C (L)} & 2, 4, 8, 16 & 4, 7, 10, 14, 25 & Head tilt-down avoiding eye-contact, head shaking and frowning.\\
			&\textbf{E (L)} & 1, 4, 10, 16 & 4, 7, 10, 14, 17 & Tilt-up, head shaking and frowning.  \\ 
			&\textbf{A (L)} & 1, 8, 9, 16 & 4, 14, 17, 25, 26 & Frequent head movements and frowning.\\
			&\textbf{N (L)} & 1, 5, 12, 16 & 4, 7, 10, 14, 25 & Few head movements, head shaking and frowning.\\ \hline
			 \multirow{8}{*}{\textbf{MIT}} & \textbf{RH (H)} & \textbf{16}, \textbf{14}, 3, 4  & 5, 10, 12, 14, 25 & Head nodding and smiling, and being expressive.\\
      &\textbf{Ex (H)} & \textbf{14}, 3, 4, 9   & 5, 10, 12, 14, 25 & Head nodding and exhibiting persistent head motion. Smiling and being expressive.\\
      &\textbf{EC (H)} & \textbf{14}, 12, 4, 5  & 6, 7, 10, 14, 25   & Head up, nodding and showing limited facial emotions.\\
      &\textbf{Fr (H)} & \textbf{16}, 3, \textbf{11}, \textbf{14} & 5, 10, 12, 14, 25 &  Frequent head movements and smiling.\\ 
			&\textbf{RH (L)} & \textbf{11}, 1, 2, 5   & 6, 7, 12, 14, 25  & Head shaking and exhibiting minimal facial expressions.\\
      &\textbf{Ex (L)} & \textbf{11}, \textbf{16}, 2, 3  & 4, 6, 7, 14, 25   &  Head shaking and nodding. Frowning and showing minimal facial expressions. \\
      &\textbf{EC (L)} & 13, 7, \textbf{16}, \textbf{11} & 6, 7, 10, 12, 25   & Frequent nodding is perceived as avoiding eye-contact.\\
      &\textbf{Fr (L)} &  3, \textbf{11}, 4, 9  & 1, 4, 6, 7, 25 & Head shaking, frowning and otherwise being minimally expressive.\\ 
      \bottomrule
    \end{tabular}
    \label{tab:explainableKineme}\vspace{-2mm}
\end{table*}

To examine kineme and AU-based explanations, we considered the top and bottom 10 percentile videos for each trait and computed the most frequent kinemes and AUs in those videos. Table~\ref{tab:explainableKineme} lists the frequently occurring kinemes and AUs for the high (H) and low (L)-rated OCEAN and Interview trait videos (top 4 kinemes based on frequency and top 5 dominant AUs are shown), and inferences thereof. We make the following observations from Table~\ref{tab:explainableKineme}:

\begin{itemize}
\item Kineme 16 is common for all OCEAN traits, while kineme 10 denoting a head-nod is seen in high OCEAN trait videos. Therefore, head-nodding in general evokes positive personality trait impressions. Likewise, AUs 25 and 26 typical of talking behavior are noted for all videos. 
\item Focusing on other kinemes, high Openness is characterized by kinemes 2 and 8, which signify persistent head movements. This finding is echoed in~\cite{KOPPENSTEINER2013}, where large motion variations are found to associate with high O impressions. Presence of AUs 12 and 14 indicates that a smiling demeanor characterizes high O. Conversely, kineme 6 denoting minimal head motion and AUs 4 and 17 typical of frowning and diffident behavior are commonly noted for low O videos.
\item Kineme 1 denoting an upward head tilt is associated with high C, while kinemes 2 and 4 depicting tilt-down and head-shaking are associated with low C. This indicates that attempting to maintain eye-contact conveys diligence and honesty, while avoiding eye-contact conveys insincerity. 
\item Extraversion appears to be conveyed better by AUs than kinemes; Dominant AUs for high E include 10, 12 and 17 indicating a friendly and talkative nature, while dominant kinemes 2 and 14 convey significant head movements. Conversely, low E is associated with kineme 4 denoting head-shaking and AUs 4, 7 and 17 indicating frowning, overall conveying a socially distant nature.
\item High Agreeableness is characterized by kineme 3 (head-nod), and AUs 12 and 14 which constitute a smile. Conversely, kinemes 1, 8 and 9 dominate low A, and they collectively convey persistent head motion. Also, AUs dominant for low A are 4, 14 and 17, cumulatively describing a frown; overall, nodding and smiling is viewed as courteous, while frequent head movements and frowning convey hostility.  
\item Emotional stability (high N) is associated with kinemes 2 and 8, and AUs 7, 12 and 17, indicating persistent head motion and facial expressiveness. On the other hand, a neurotic trait is conveyed via limited head motion and head-shaking (kinemes 1, 5, 12) and frowning (described by AUs 4, 7, 10).
\item While kinemes for the MIT videos are less discernible, due to smaller face size (Fig.~\ref{fig:Ov_FICS_MIT}) and the fact that they capture an interactional setting, some patterns are nevertheless evident as seen in Fig.~\ref{fig:selectKinemes_MIT}; these kinemes are highlighted in Table~\ref{tab:explainableKineme}. As with FICS, Kineme 14 denoting a head-nod is commonly observed for all high trait videos, while kineme 11 depicting a head-shake is common for all low-trait videos.  
\item High RH scores are elicited with expressive facial behavior involving head-nodding and smiling.   Conversely, low RH scores are associated with head-shaking and exhibiting limited facial expressions. Highly excited behavior is associated with identical AUs as high RH, and persistent head motion. Inversely, low excitement scores are connected with head shaking, and limited facial emotions. 
\item Identical AUs are observed for both high and low eye-contact, implying that head movements primarily impact eye-contact impressions. Head nodding (kineme 14) is associated with high EC, while kinemes 11 and 16 depicting head shaking and frequent head-nodding elicit low EC scores. Therefore interestingly, while head nodding is beneficial, frequent nodding is perceived as avoiding eye-contact. 
\item High friendliness is characterized by kinemes 11, 14 and 16, signifying persistent head motion along with expressive and smiling facial movements (AUs 5, 12 and 14). Conversely, low friendliness is associated with head-shaking (kineme 11) and frowning (AUs 4, 6, 7).                                              
\end{itemize}

Overall, kineme and AU patterns offer intuitive explanations for the considered personality and interview traits. In addition, they enable efficient trait prediction as described in the next section.

\vspace{-3.5mm}

\section{Experiments and Results}\label{Sec:ER}

\subsection{Datasets}\label{sec:datasets}

\begin{sloppypar}
\noindent \textbf{FICS:} The FICS dataset~\cite{escalante2020modeling} comprises 10K ($\approx$15s long) \emph{YouTube} self-presentation videos, and is labeled via crowdworkers for the OCEAN personality traits. As Neuroticism is a negative trait, N scores encode inverse of Neuroticism scores in this dataset. FICS contains 6K \emph{training}, 2K \emph{validation} and 2K \emph{test} videos. 	\\
\end{sloppypar}
\noindent  \textbf{MIT:} The MIT interview~\cite{Naim18} dataset comprises 138 mock internship interview videos, with an average length of 4.7 minutes. These videos were rated for 16 traits by crowdworkers; along with the four traits considered in Sec.~\ref{Sec:UC}, we also evaluate prediction performance on the \emph{overall} interview score (Ov) in this section.  

\subsection{Experimental settings}\label{Sec:ES}

\textbf{Prediction Type
:} For our experiments, we modeled the personality and interview trait scores as either continuous or discrete variables (upon thresholding trait scores at their median value), and present regression (Tables~\ref{tab:FICS_reg},~\ref{tab:MIT_reg}) and classification (Tables~\ref{tab:FICS_class},~\ref{tab:MIT_class}) results. As the FICS dataset is already partitioned into the train, validation and test sets, we employed the validation set for fine-tuning model parameters, or for early stopping of model training for LSTM. On the other hand, as the MIT dataset only comprises 138 videos, we report performance in the form of $\mu \pm \sigma$ over five repetitions of 10-fold video cross validation (total of 50 runs) in Tables~\ref{tab:MIT_reg} and~\ref{tab:MIT_class}. Validation sets were obtained by randomly holding out 10\% of the training data during each run. \\

\noindent \textbf{Chunk vs video-level predictions:} In line with the thin-slice approach for behavioral trait prediction, we segmented the original videos into smaller chunks, 3, 5 and 7s for the FICS data and 5--60s chunks for the MIT data (see Fig.~\ref{fig:Chunk_vs_Vid_MIT}) and repeated the video label for all chunks. We then computed metrics over a) all chunks (chunk-level performance), and b) over all videos by assigning the majority label over all chunks to each video (video-level performance).  

\subsection{Performance Metrics} For regression, we considered the accuracy (Acc) and Pearson Correlation Coefficient (PCC) metrics. As in~\cite{Gucluturk2018}, Acc is measured as 1-MAE, the mean absolute error in the predictions with respect to groud-truth scores; the PCC measures how well the predictions correlate with the ground-truth scores (PCC=1 for precise predictions). For classification, we employed the accuracy (Acc) and F1-score metrics. Given the imbalanced class distributions for some traits (see Table~\ref{tab:class-dist}), the F1-score denoting the harmonic mean of precision and recall is more suited for performance evaluation. 	

\subsection{Models} 

\noindent \textbf{Linear regression with principal components (PCA-Lin-Reg):} We collated the yaw, pitch and roll values over each chunk, and then performed principal component analysis (PCA) over the training data. As many principal components explaining 90\% data variance were preserved, and linear regression was performed on the same. This method ignores the temporal head and facial motion dynamics. \\

\noindent \textbf{Hidden Markov Model for classification (HMM Kin):} To learn trait-wise temporal head motion patterns, we employed a HMM to iteratively deduce model parameters and class labels from kineme sequences via the Baum-Welch algorithm. \\

\begin{figure}[!htb]
      \centering
     \includegraphics[width=\linewidth]{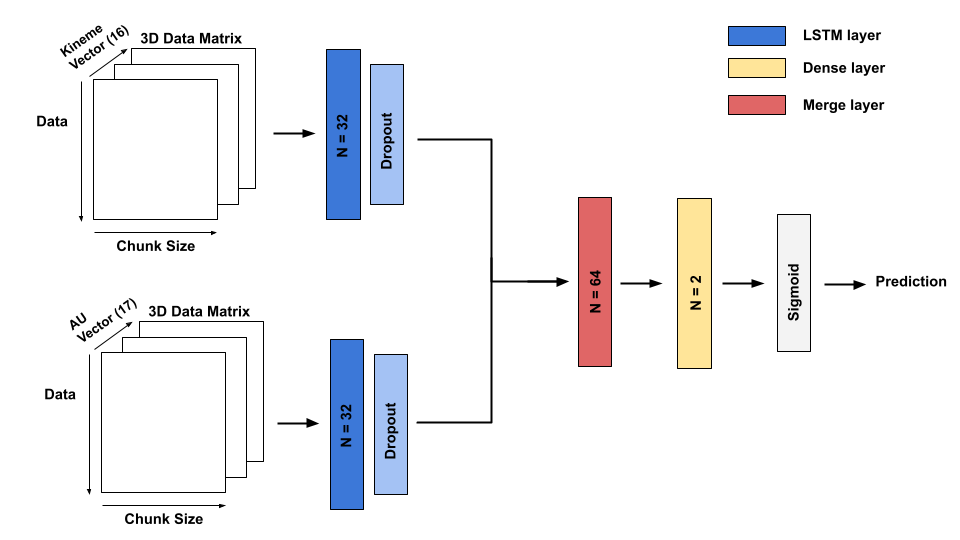}\vspace{-2mm}
    \caption{Kineme$+$AU LSTM classification architecture. The dense layer output involves a single neuron with linear activation for regression. Best viewed in color and zoom.} \label{fig:Kin_AU_arch}\vspace{-4mm}
\end{figure}

\noindent \textbf{Long short-term memory for regression and classification:} LSTMs denote another efficient methodology to learn latent temporal head and facial motion dynamics. We trained LSTMs with the kineme sequences \textbf{(LSTM Kin)}, AU sequences \textbf{(LSTM AU)} and their combination \textbf{(LSTM Kin+AU FF)}. The kineme sequences were input in the one-hot encoding form, where the kineme corresponding to a given time-window is coded to 1 and the other 15 coded to 0. AUs on the other hand were input in a coded vector form, where all AUs deemed dominant upon thresholding within a time-window were set to 1 and others set to 0. To fuse the kineme and AU-based representations (denoted as feature fusion or FF), we employed the architecture shown in Fig.~\ref{fig:Kin_AU_arch}.  Kineme features form a 3D matrix of 16 (one-hot-kineme-vector) $\times$ data-points $\times$ chunk-length, while the AU matrix is of size 17 (vector of active AUs coded to 1) $\times$ data-points $\times$ chunk-length. A single hidden LSTM layer with 32 neurons was employed for kinemes and AUs, while these outputs were merged to obtain 64 neurons for FF. The LSTM layer is followed by a dense layer involving two neurons with sigmoidal activation for classification, and one neuron with linear activation for regression. In both cases, a dropout value of 0.2 was employed for the LSTM layer to prevent overfitting, and an Adam optimizer with learning rate of 0.01 was utilized for training. Binary-cross entropy was defined as the loss function for classification, and mean absolute error for regression. \\

\noindent \textbf{2D CNN for regression and classification:} A 19 layered VGG model, which processes images (video frames) was used. Upon removing the output layer, two hidden dense layers with 512 and 64 neurons respectively were added along with output layer involving 5 neurons (one neuron
each for the OCEAN/Interview traits). Mean squared error (MSE) for regression, and binary cross-entropy (BCE) loss for classification were used to train on a random frame from each video, with learning rate of 1e-4 and a batch size of 64. \\

\begin{sloppypar}
\noindent \textbf{Decision fusion for regression and classification {(LSTM Kin+AU (DF))}:} Apart from LSTM-based kineme AU fusion, we attempted fusion of the unimodal LSTM predictions. We adopted the fusion weight estimation approach proposed in~\cite{KOELSTRA2013164}. Assuming that the test sample score is $\alpha p_{Kin} + (1-\alpha) p_{AU}, \alpha \in [0,1]$, where $p_{Kin}$ and $p_{AU}$ are the individual classifier (regressor) scores, we performed grid-search incrementing $\alpha$ in steps of 0.01 to estimate the optimal $\alpha^*$ maximizing PCC for regression and F1-score for classification; $\alpha^*$ was then applied to compute test sample scores. 
\end{sloppypar}

\begin{table*}[!htbp]
\centering
		\small
		\caption{Trait-wise train (Tr) and test (Te) class distributions for the FICS and MIT datasets obtained for classification experiments. MIT class distributions correspond to 1-minute video samples employed for analysis.}\label{tab:class-dist} \vspace{-2mm}
		\begin{tabular}{|c|cc|cc|cc|cc|cc|cc|cc|cc|cc|cc|}
		\toprule
		& \multicolumn{10}{|c|}{\textbf{FICS}} & \multicolumn{10}{|c|}{\textbf{MIT}} \\ \hline
		& \multicolumn{2}{|c|}{\textbf{O}} & \multicolumn{2}{|c|}{\textbf{C}} & \multicolumn{2}{|c|}{\textbf{E}} & \multicolumn{2}{|c|}{\textbf{A}} & \multicolumn{2}{|c|}{\textbf{N}} & \multicolumn{2}{|c|}{\textbf{Ov}} & \multicolumn{2}{|c|}{\textbf{RH}} & \multicolumn{2}{|c|}{\textbf{Ex}} & \multicolumn{2}{|c|}{\textbf{EC}} & \multicolumn{2}{|c|}{\textbf{Fr}}\\ \hline
		\textbf{Label} & \bf Tr & \bf Te & \bf Tr & \bf Te & \bf Tr & \bf Te & \bf Tr & \bf Te & \bf Tr & \bf Te & \bf Tr & \bf Te & \bf Tr & \bf Te & \bf Tr & \bf Te & \bf Tr & \bf Te & \bf Tr & \bf Te \\
		\textbf{-ve} & 0.53 & 0.52 & 0.51 & 0.51 & 0.51 & 0.51 & 0.53 & 0.53 & 0.52 & 0.52 & 0.50 & 0.50 & 0.59 & 0.59 & 0.50 & 0.50 & 0.43 & 0.43 & 0.39 & 0.39\\
		\textbf{+ve} & 0.47 & 0.48 & 0.49 & 0.49 & 0.49 & 0.49 & 0.47 & 0.47 & 0.48 & 0.48 & 0.50 & 0.50 & 0.41 & 0.41 & 0.50 & 0.50 & 0.57 & 0.57 & 0.61 & 0.61\\
		 \bottomrule
    \end{tabular}
    \vspace{-2mm}
\end{table*}

 \begin{table*}[!htbp]
    \centering
		\small
    \caption{FICS Regression results: Accuracy and PCC values for different methods are tabulated.} \vspace{-2mm}
    \begin{tabular}{|l|cc|cc|cc|cc|cc|cc|}
    \toprule
     \bf Trait & \multicolumn{2}{|c|}{\textbf{PCA Lin-Reg}}  & \multicolumn{2}{|c|}{\textbf{LSTM Kin}} & \multicolumn{2}{|c|}{\textbf{LSTM AU}} & \multicolumn{2}{|c|}{\textbf{LSTM Kin+AU (FF)}}  & \multicolumn{2}{|c|}{\textbf{LSTM Kin+AU (DF)}}  & \multicolumn{2}{|c|}{\textbf{2D-CNN}}\\
		& \textbf{Acc} & \textbf{PCC}  & \textbf{Acc} & \textbf{PCC} & \textbf{Acc} & \textbf{PCC} & \textbf{Acc} & \textbf{PCC} & \textbf{Acc} & \textbf{PCC} & \textbf{Acc} & \textbf{PCC}\\
     \hline
      \textbf{Open}   &  0.884 & 0.085 & 0.872 & 0.060 & 0.889 & 0.370 & 0.892 & 0.368 & 0.893 & 0.382 &\textbf{0.906} & \textbf{0.392}\\
			\textbf{Con}    &  0.875 & 0.086 & 0.864 & 0.027 & 0.882 & 0.317 & 0.880 & 0.304 & 0.882 & 0.282 &\textbf{0.908} & \textbf{0.295}\\
			\textbf{Extra}  &  0.877 & 0.060 & 0.869 & 0.048 & 0.891 & 0.491 & 0.893 & 0.474 & 0.891 & 0.485 &\textbf{0.907} & \textbf{0.492}\\
			\textbf{Agree}  &  0.892 & 0.035 & 0.885 & 0.046 & 0.897 & 0.251 & 0.892 & 0.253 & 0.896 & 0.275 &\textbf{0.906} & \textbf{0.283}\\
			\textbf{Neuro}  &  0.877 & 0.071 & 0.867 & 0.051 & 0.885 & 0.370 & 0.884 & 0.365 & 0.887 & 0.387 &\textbf{0.903} & \textbf{0.395}\\
			
      \bottomrule
    \end{tabular}
    \label{tab:FICS_reg}\vspace{-2mm}
\end{table*}

\begin{table*}[!htbp]
    \centering
		 \fontsize{7}{7}\selectfont
		\renewcommand{\arraystretch}{1.5}
    \caption{MIT Regression results: Accuracy and PCC values are tabulated.} \vspace{-2mm}
    \begin{tabular}{|l|cc|cc|cc|cc|cc|cc|}
    \toprule
     \bf Trait &  \multicolumn{2}{|c|}{\textbf{PCA Lin-Reg}}  &  \multicolumn{2}{|c|}{\textbf{LSTM Kin}} &  \multicolumn{2}{|c|}{\textbf{LSTM AU}} & \multicolumn{2}{|c|}{\textbf{LSTM Kin+AU (FF)}} & \multicolumn{2}{|c|}{\textbf{LSTM Kin+AU (DF)}} & \multicolumn{2}{|c|}{\textbf{2D-CNN}}\\
		 & \textbf{Acc} & \textbf{PCC}  & \textbf{Acc} & \textbf{PCC} & \textbf{Acc} & \textbf{PCC} & \textbf{Acc} & \textbf{PCC} & \textbf{Acc} & \textbf{PCC} & \textbf{Acc} & \textbf{PCC}\\
     \hline
      \textbf{Ov}   & 0.86$\pm$0.00  & 0.15$\pm$0.26 & 0.93$\pm$0.04 & 0.84$\pm$0.26 & 0.93$\pm$0.04 & 0.84$\pm$0.26 & \textbf{0.97$\pm$0.03} &\textbf{ 0.92$\pm$0.17} &  0.95$\pm$0.04 & 0.89$\pm$0.21 & 0.86$\pm$0.03 & 0.47±0.25\\
			\textbf{RH}    & 0.83$\pm$0.03  & 0.13$\pm$0.26 & \textbf{0.95$\pm$0.03} & \textbf{0.93$\pm$0.10} & \textbf{0.95$\pm$0.03} & \textbf{0.93$\pm$0.10} & 0.96$\pm$0.04 & 0.91$\pm$0.21 &  0.94$\pm$0.04 & 0.90$\pm$0.19 &0.85$\pm$0.03 & 0.45$\pm$0.24\\
			\textbf{Ex}   & 0.82$\pm$0.04  & 0.05$\pm$0.20 & 0.94$\pm$0.04 & 0.89$\pm$0.20 & 0.94$\pm$0.04 & 0.89$\pm$0.20 & \textbf{0.96$\pm$0.05} & \textbf{0.91$\pm$0.20} &  0.94$\pm$0.04 & 0.91$\pm$0.17 & 0.85$\pm$0.03 & 0.57$\pm$0.18\\
			\textbf{EC}    & 0.83$\pm$0.03  & 0.10$\pm$0.21 & 0.94$\pm$0.04 & 0.89$\pm$0.13 & 0.94$\pm$0.04 & 0.89$\pm$0.22 & \textbf{0.96$\pm$0.04} & \textbf{0.94$\pm$0.13} &  0.95$\pm$0.04 & 0.91$\pm$0.16 & 0.84$\pm$0.03 & 0.44$\pm$0.22\\
			\textbf{Fr}  & 0.82$\pm$0.04  & 0.02$\pm$0.28 & 0.95$\pm$0.03 & 0.93$\pm$0.10 & 0.95$\pm$0.03 & 0.93$\pm$0.10 & \textbf{0.97$\pm$0.03} & \textbf{0.96$\pm$0.08} &  0.95$\pm$0.03 & 0.94$\pm$0.08 & 0.86$\pm$0.03 & 0.58 $\pm$0.20\\
			
      \bottomrule
    \end{tabular}
    \label{tab:MIT_reg}
    \vspace{-2mm}
\end{table*}

\begin{table*}[!htbp]
    \centering
		\small
    \caption{FICS Classification results: Accuracy and F1 values are tabulated.} \vspace{-2mm}
    \begin{tabular}{|l|cc|cc|cc|cc|cc|cc|}
    \toprule
     \bf Trait & \multicolumn{2}{|c|}{\textbf{HMM Kin}}  &  \multicolumn{2}{|c|}{\textbf{LSTM Kin}} &  \multicolumn{2}{|c|}{\textbf{LSTM AU}} & \multicolumn{2}{|c|}{\textbf{LSTM Kin+AU (FF)}} & \multicolumn{2}{|c|}{\textbf{LSTM Kin+AU (DF)}} & \multicolumn{2}{|c|}{\textbf{2D-CNN}}\\     \hline
		 & \textbf{Acc} & \textbf{F1}  & \textbf{Acc} & \textbf{F1} & \textbf{Acc} & \textbf{F1} & \textbf{Acc} & \textbf{F1} & \textbf{Acc} & \textbf{F1} & \textbf{Acc} & \textbf{F1}\\
      \textbf{Open}   &  0.529 & 0.505 & 0.519 & 0.516 & 0.635 & 0.634 & 0.629 & 0.628 & 0.632 & 0.632 & 0.697 & \textbf{0.697}\\
			\textbf{Con}    &  0.524 & 0.515 & 0.513 & 0.513 & 0.618 & 0.618 & 0.604 & 0.604 & 0.599 & 0.599 & 0.705 & \textbf{0.705}\\
			\textbf{Extra}  &  0.514 & 0.509 & 0.505 & 0.505 & 0.651 & 0.651 & 0.648 & 0.648 & 0.657 & 0.653 & 0.712 & \textbf{0.712} \\
			\textbf{Agree}  &  0.528 & 0.497 & 0.481 & 0.479 & 0.580 & 0.580 & 0.593 & 0.586 & 0.584 & 0.583 & 0.656 & \textbf{0.656}\\
			\textbf{Neuro}  &  0.524 & 0.506 & 0.523 & 0.518 & 0.627 & 0.624 & 0.626 & 0.623 & 0.620 & 0.616 & 0.717 & \textbf{0.717}\\
			
      \bottomrule
    \end{tabular}
    \label{tab:FICS_class}\vspace{-2mm}
\end{table*}

\begin{table*}[!htbp]
    \centering
		 \fontsize{7}{7}\selectfont
		\renewcommand{\arraystretch}{1.5}
    \caption{MIT Classification results: Accuracy and F1 values are tabulated.} \vspace{-2mm}
    \begin{tabular}{|l|cc|cc|cc|cc|cc|cc|}
    \toprule
     \bf Trait &  \multicolumn{2}{|c|}{\textbf{HMM}}  &  \multicolumn{2}{|c|}{\textbf{LSTM Kin}} &  \multicolumn{2}{|c|}{\textbf{LSTM AU}} & \multicolumn{2}{|c|}{\textbf{LSTM Kin+AU (FF)}} & \multicolumn{2}{|c|}{\textbf{LSTM Kin+AU (DF)}} & \multicolumn{2}{|c|}{\textbf{2D-CNN}}\\
		 & \textbf{Acc} & \textbf{F1}  & \textbf{Acc} & \textbf{F1} & \textbf{Acc} & \textbf{F1} & \textbf{Acc} & \textbf{F1} & \textbf{Acc} & \textbf{F1} & \textbf{Acc} & \textbf{F1}\\
     \hline
      \textbf{Ov}   & 0.56$\pm$0.17 & 0.48$\pm$0.18 & 0.83$\pm$0.11 & 0.82$\pm$0.13 & 0.82$\pm$0.14 & 0.81$\pm$0.15 & 0.80$\pm$0.14 & 0.80$\pm$0.14 & 0.85$\pm$0.13 & \textbf{0.85$\pm$0.14} & 0.65$\pm$0.09 & 0.64$\pm$0.08\\
			\textbf{RH}   & 0.55$\pm$0.14 & 0.34$\pm$0.27 & 0.79$\pm$0.12 & 0.79$\pm$0.12 & 0.83$\pm$0.13 & 0.83$\pm$0.14 & 0.81$\pm$0.12 & 0.80$\pm$0.12 & 0.84$\pm$0.11 & \textbf{0.83$\pm$0.12} & 0.59$\pm$0.07 & 0.58$\pm$0.07\\
			\textbf{Ex}   & 0.57$\pm$0.14 & 0.51$\pm$0.10 & 0.82$\pm$0.13 & 0.82$\pm$0.13 & {0.82$\pm$0.12} & 0.82$\pm$0.10 & 0.79$\pm$0.13 & 0.79$\pm$0.13 & 0.83$\pm$0.11 & \textbf{0.82$\pm$0.12} & 0.72$\pm$0.08 & 0.71$\pm$0.08\\
			\textbf{EC}  & 0.53$\pm$0.17 & 0.44$\pm$0.24 & 0.79$\pm$0.13 & 0.79$\pm$0.13 & 0.81$\pm$0.12 & 0.80$\pm$0.13 & 0.78$\pm$0.12 & 0.76$\pm$0.13 & 0.84$\pm$0.13 & \textbf{0.83$\pm$0.14} & 0.57$\pm$0.07 & 0.55$\pm$0.09\\
			\textbf{Fr}   & 0.57$\pm$0.18 & 0.54$\pm$0.12 &  0.80$\pm$0.15 & 0.80$\pm$0.16 & 0.86$\pm$0.09 & 0.85$\pm$0.09 & 0.84$\pm$0.10 & 0.84$\pm$0.11 & 0.87$\pm$0.11 & \textbf{0.86$\pm$0.12} & 0.61$\pm$0.08 & 0.61$\pm$0.08\\
			
      \bottomrule
    \end{tabular}
    \label{tab:MIT_class}\vspace{-4mm}
\end{table*}
\begin{figure}[!htbp]
\centering
 \includegraphics[width=\linewidth]{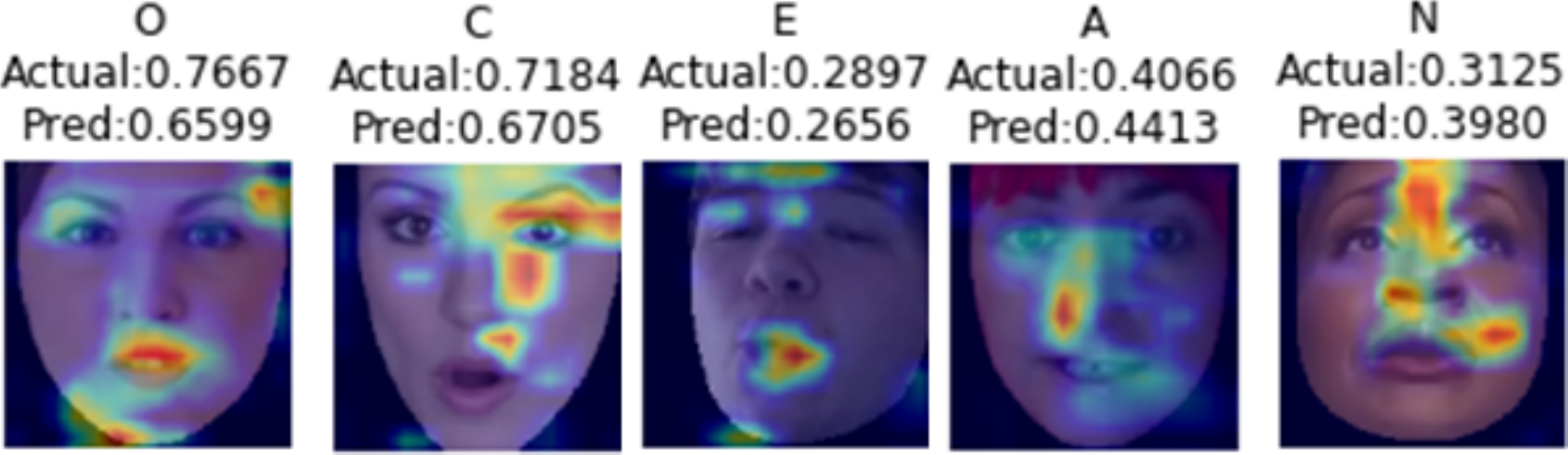}
\vspace{-5mm}\caption{Exemplar Gradcam~\cite{Selvaraju_2019} maps for CNN predictions.}\label{fig:Gradcam_out}
    \vspace{-6mm}
\end{figure}
\begin{figure*}[!htbp]
\centering
    \includegraphics[width=0.48\linewidth]{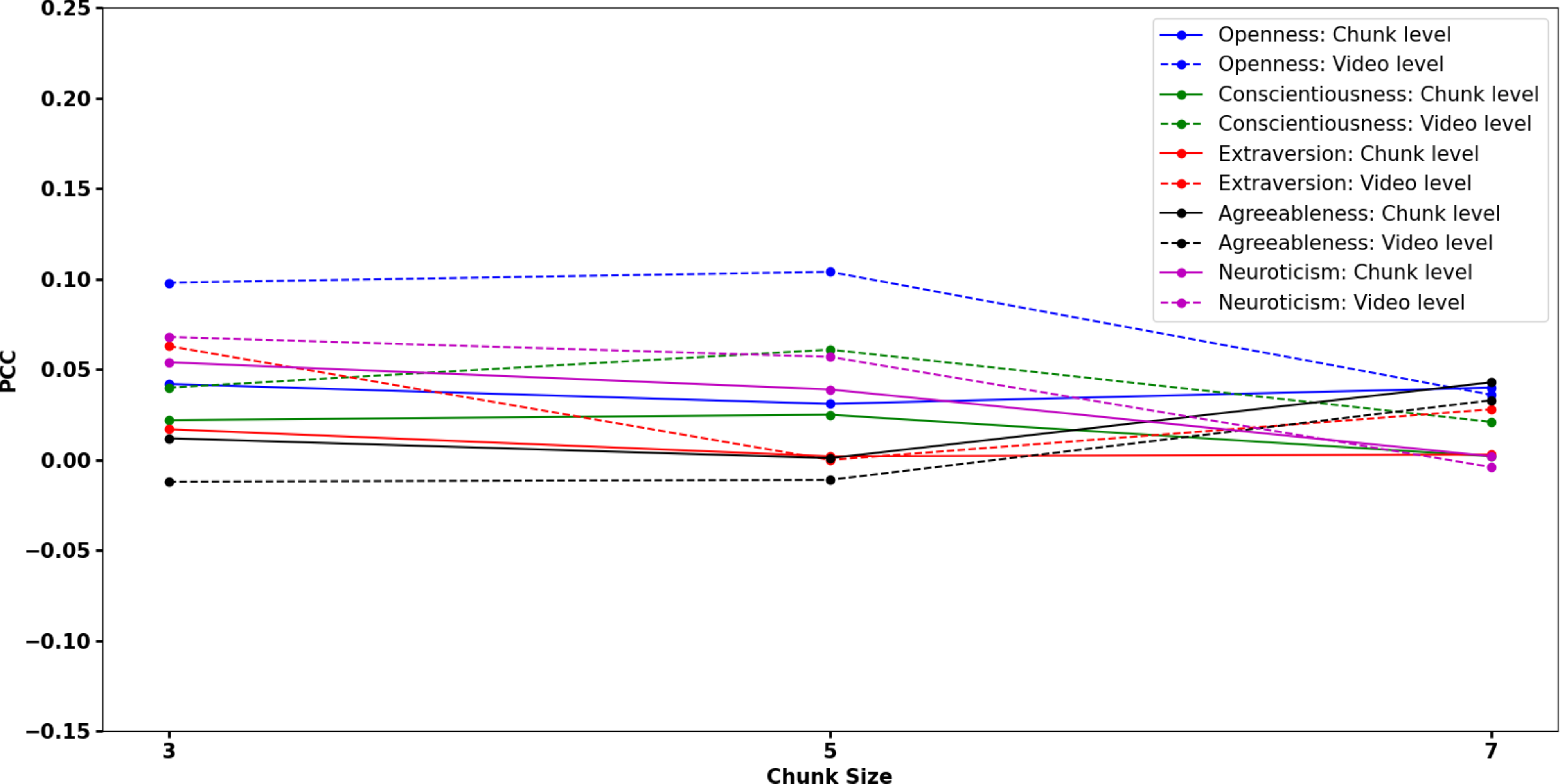}\hspace{0.05cm}\includegraphics[width=0.48\linewidth]{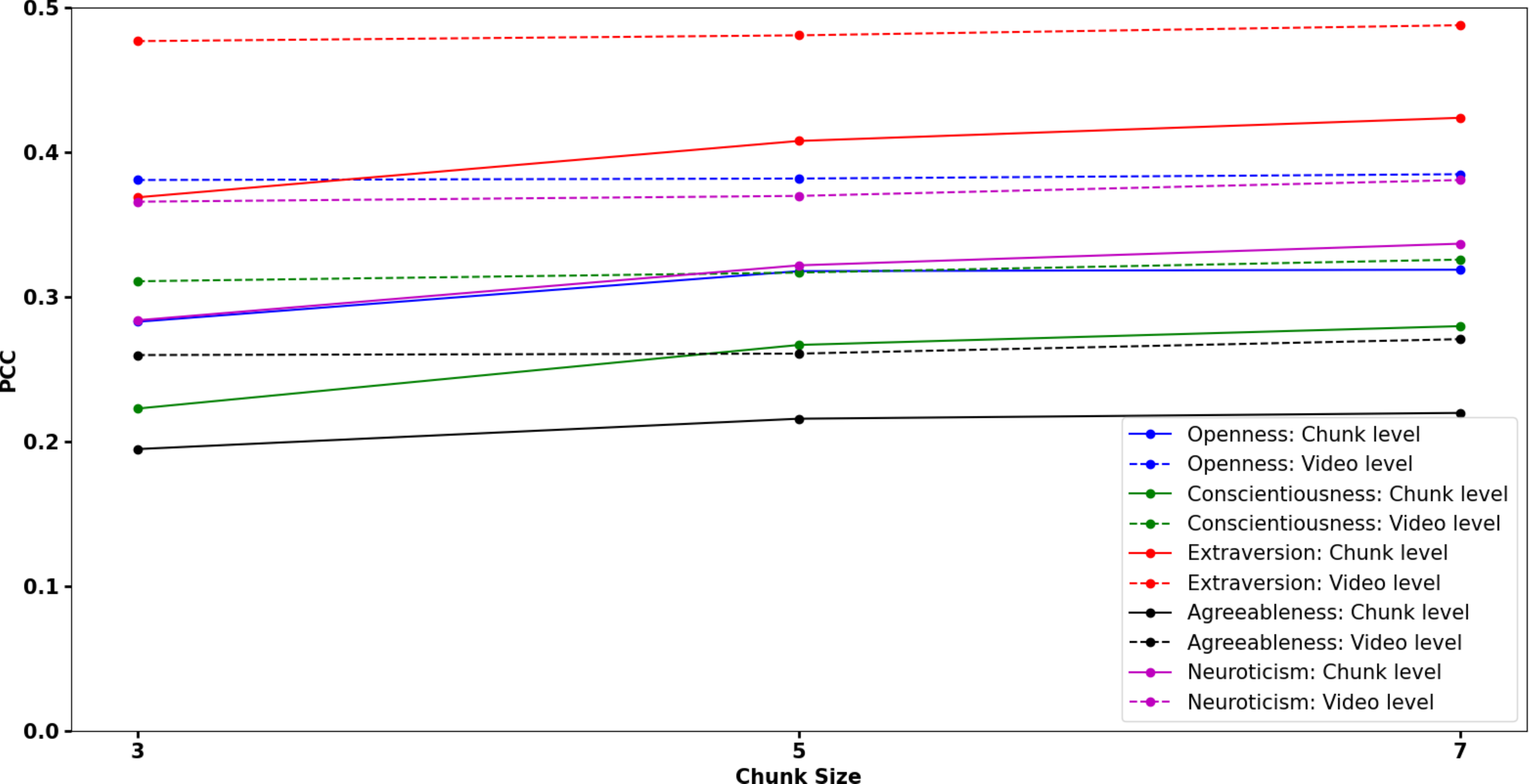}
        \vspace{-2mm}\caption{Chunk vs video-level predictions with kinemes (left) and AUs (right) for the FICS dataset.}\label{fig:Chunk_vs_Vid_FICS}
    \vspace{.5mm}
        \centering
        \includegraphics[width=0.48\linewidth,height=5cm]{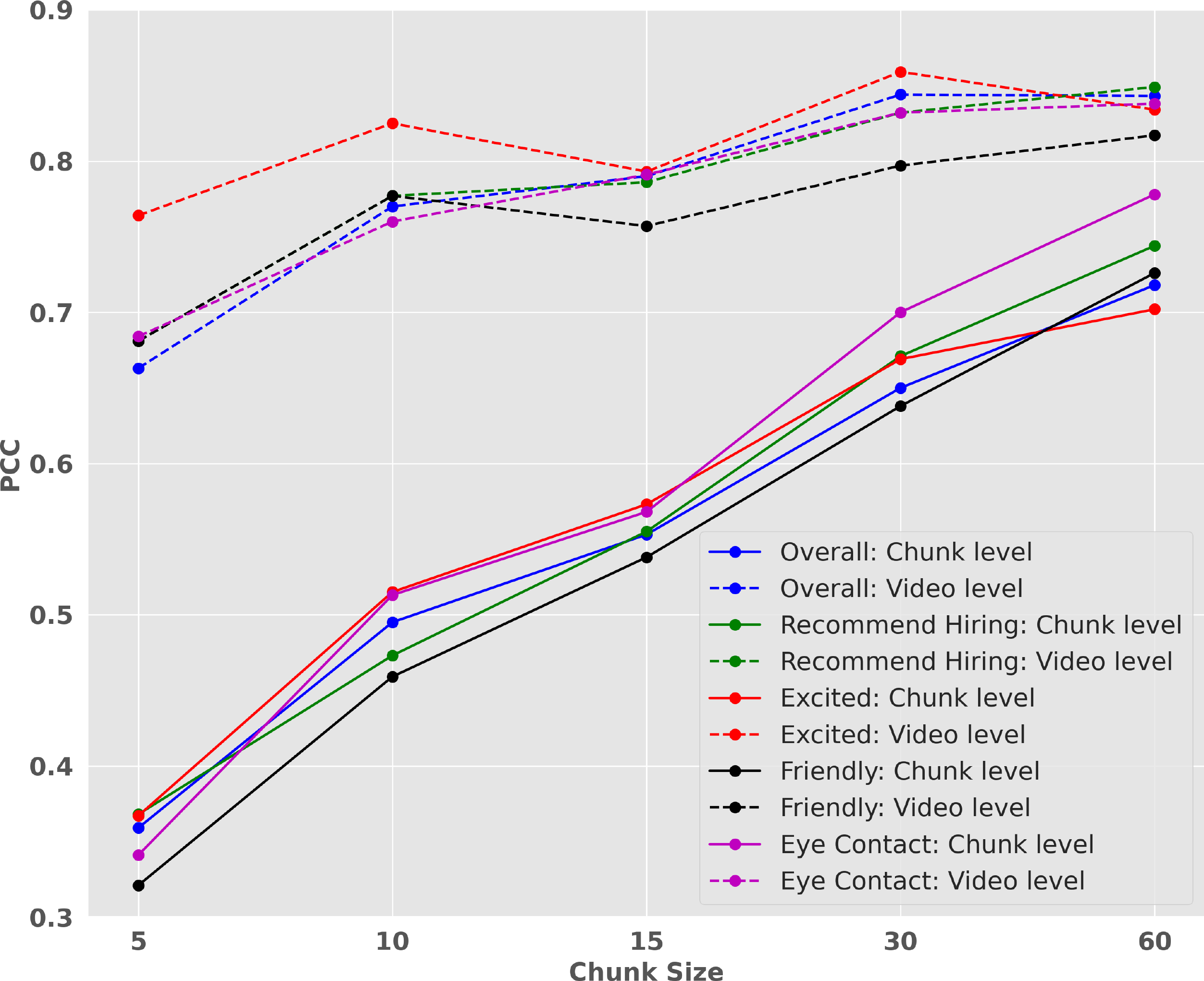}\hspace{0.05cm}\includegraphics[width=0.48\linewidth,height=5cm]{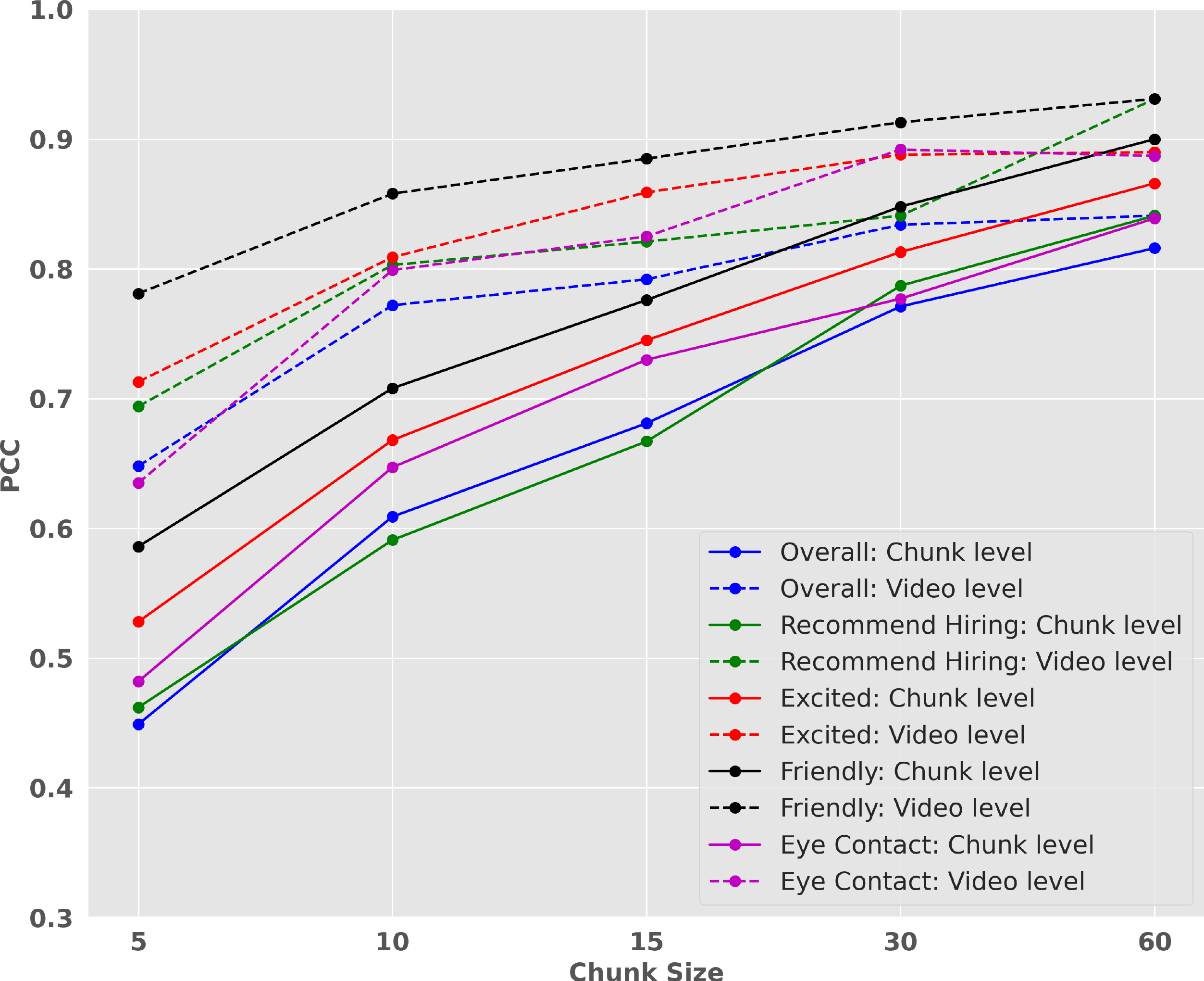}
        \vspace{-2mm}\caption{Chunk vs video-level predictions with kinemes (left) and AUs (right) for the MIT dataset.}\label{fig:Chunk_vs_Vid_MIT}\vspace{-4mm}
\end{figure*}

\subsection{Results and Discussion}
Tables~\ref{tab:FICS_reg} and~\ref{tab:MIT_reg} present regression results, while Tables~\ref{tab:FICS_class},~\ref{tab:MIT_class} tabulate classification results. We make the following remarks therefrom:
\begin{itemize}
\item One can note from Tables~\ref{tab:FICS_reg} and~\ref{tab:MIT_reg} that very low PCC values are noted even for relatively high accuracies for the FICS dataset; these results convey that larger variability in prediction scores is observed for the FICS videos, and that PCC is more suited for MIT performance evaluation. Also, comparing Tables~\ref{tab:FICS_reg},~\ref{tab:MIT_reg} with Tables~\ref{tab:FICS_class},~\ref{tab:MIT_class}, much higher performance is achieved for regression than classification. This can be attributed to the Gaussian-distributed FICS OCEAN trait scores around 0.5 (see Figs 12 and 13 in~\cite{escalante2020modeling}), and the MIT Interview trait scores about the median.
\item Focusing on Table~\ref{tab:FICS_reg}, PCA Lin-Reg which ignores temporal head-motion dynamics, performs better than LSTM Kin; this implies that the extracted kinemes cannot adequately describe the FICS OCEAN scores. Very poor PCC scores are obtained with both PCA Lin-Reg and LSTM Kin. 
\item AUs achieve superior LSTM-based prediction than kinemes. However, complementarity of the kineme and AU encodings in general enables slightly superior performance with both feature and decision fusion, with both fusion schemes performing comparably. 2D-CNN achieves optimal predictions on the FICS videos; the explanatory power of CNNs is however limited. Grad-cam visualizations (Fig.~\ref{fig:Gradcam_out}) which highlight regions deemed critical for model prediction focus on the eye, mouth and nasal regions as in~\cite{Ventura17}. Unlike kinemes however, trait-characteristic temporal dynamics of these regions are not conveyed by these saliency maps.
\item Kinemes however demonstrate superior predictive power for interview traits (Table~\ref{tab:MIT_reg}), considerably outperforming Lin-Reg and performing identical to AUs. Fusing unimodal features/decisions is beneficial for the MIT dataset, with feature fusion outperforming decision fusion. 
\item Comparing against the Resnet-based trait prediction model proposed in~\cite{Gucluturk2018}, we note that the 2D-CNN achieves similar or superior performance, while the Kineme/AU-based LSTM performs inferiorly.
\item On the smaller MIT dataset however, LSTM Kin considerably outperforms 2D-CNN, revealing the need for large training data to effectively tune VGG-type networks. 
\item In Table~\ref{tab:FICS_reg}, 2D-CNN achieves highest accuracy for  Conscientiousness, while Agreeableness is predicted best via all other methods. Highest PCC values are achieved for Friendliness with all methods excepting PCA Lin-Reg in Table~\ref{tab:MIT_reg}.
\item Focusing on Table~\ref{tab:FICS_class}, HMM Kin and LSTM Kin perform worst for OCEAN trait prediction. LSTM AU performs considerably better than LSTM Kin and similar to the feature and decision fusion schemes, implying that multimodal fusion is not very beneficial for the FICS dataset. The 2D-CNN achieves optimal trait classification, substantially outperforming other methods. With all schemes, least classification performance is observed for Agreeableness, while Extraversion is best classified by all schemes excepting HMM Kin and LSTM Kin.   
\item With respect to Table~\ref{tab:MIT_class}, the LSTM network employing kineme features achieves considerably higher classification performance than HMM Kin. This trend again confirms that kinemes are effective predictors of interview traits. LSTM AU performs slightly better than LSTM Kin, while the decision fusion framework achieves optimal classification, marginally outperforming LSTM AU. 
\item As with regression, LSTM Kin achieves substantially better F1-scores than 2D-CNN, confirming the efficacy of the kineme representation with fewer training data. The Friendliness trait is best isolated by four of the six classifiers.
\end{itemize}

The above results are achieved by considering 15s chunks for the FICS videos and 1-minute chunks for the MIT videos, and assigning the video label as the majority label over all chunk predictions. As mentioned in Sec.~\ref{Sec:ES}, we attempted trait prediction from thin behavioral slices (chunks) of varying lengths and evaluated prediction performance at the chunk and video levels. 

Figures~\ref{fig:Chunk_vs_Vid_FICS} and ~\ref{fig:Chunk_vs_Vid_MIT} present corresponding results. In both figures, higher PCC values are achieved with video-level labels than chunk-level labels, implying that while episodic behaviors may be inconsistent with one another, trait-specific behaviors are indeed homogeneous over a long time-span. From Fig.~\ref{fig:Chunk_vs_Vid_FICS}, we note that kineme-based trait prediction performance generally decreases with larger time-slices, while AU-based predictions become more accurate with larger time-slices. Conversely, one can note a general increase in video and chunk-level PCC values with both Kineme and AU features for the MIT dataset. Interestingly, even with 5s slices, we obtain a PCC > 0.3 at the chunk-level and PCC > 0.65 at the video level with kineme features, and still higher values are obtained with AUs. These trends suggest that AUs describing facial behavior, encode more trait-specific information than kinemes characterizing head motion, consistent with one's expectation.
\vspace{-4mm}

\section{Conclusion}\label{Sec:DC}

This work demonstrates that kinemes, denoting fundamental and interpretable head-motion units, achieve effective personality and interview trait prediction and also provide behavioral explanations for different traits consistent with prior findings. Our empirical results confirm that characteristic head gesture behaviors indeed exist for different traits, and this work contributes to the understanding of the role of head gestures in behavioral trait prediction. 

Combining facial action unit information with kinemes enables more efficient trait prediction and intuitive explanations. This work extracts kinemes over fixed time windows, which represents a limitation as head movement behaviors can vary across individuals, and also be conditioned on speaking behavior (relaxed vs animated speech).
Existence of trait-specific kineme patterns can be exploited to synthesize head motion for virtual agents with specified personality traits. The finding that apparent trait impressions are explainable in terms of head and facial motion patterns can also be utilized to developing assistive technologies to help users improve their public speaking and interactional skills. Nevertheless, the proposed behavioral analytic tools are intended to support and complement human decision-making, and the authors do not advocate the exclusive use of such technology for complex processes such as job recruitment. 
\vspace{-4mm}
\section*{Acknowledgement}
Thanks to A. Samanta, IIT Kanpur for sharing the kineme code.


\bibliographystyle{ACM-Reference-Format}
\bibliography{references}

\end{document}